\documentclass[letterpaper, 10 pt, conference]{ieeeconf}

\IEEEoverridecommandlockouts

\usepackage{graphics}
\usepackage{epsfig}
\usepackage{mathptmx}
\usepackage{times}
\usepackage{amsmath}
\usepackage{amssymb}
\usepackage{bm}
\let\mathcal\relax
\DeclareMathAlphabet{\mathcal}{OMS}{cmsy}{m}{n}
\usepackage{url}
\usepackage{booktabs}
\usepackage{array}
\usepackage{multirow}
\usepackage{makecell}
\usepackage{colortbl}
\usepackage{xcolor}
\usepackage{caption, subcaption}
\usepackage{algorithm}
\usepackage{algorithmic}
\usepackage{xspace}
\usepackage{graphicx}

\definecolor{cvprblue}{rgb}{0.21,0.49,0.74}

\makeatletter
\let\NAT@parse\undefined
\makeatother
\usepackage[pagebackref=true,breaklinks,colorlinks,citecolor=cvprblue,linkcolor=cvprblue,urlcolor=cvprblue,bookmarks=true]{hyperref}
\usepackage[capitalize]{cleveref}

\crefname{section}{Sec.}{Secs.}
\Crefname{section}{Sec.}{Secs.}
\crefname{figure}{Fig.}{Figs.}
\Crefname{figure}{Fig.}{Figs.}
\crefname{table}{Tab.}{Tabs.}
\Crefname{table}{Tab.}{Tabs.}
\crefname{algorithm}{Alg.}{Algs.}
\Crefname{algorithm}{Alg.}{Algs.}

\newcommand{\bee}{\raisebox{-0.15em}{\includegraphics[height=1.2em]{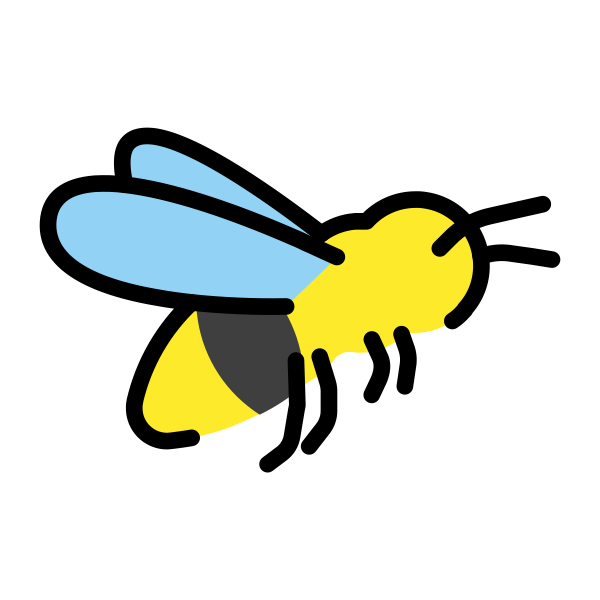}}}
\newcommand{\ours}{\mbox{\textbf{\textsc{Bee}}}}

\title{\LARGE \bf
\bee\ \ours{}: Intervention-Adaptive Real-World Reinforcement Learning with Vision-Language-Action Models
}

\author{%
\textbf{Weihui Zhao}\textsuperscript{1,2} \quad
\textbf{Xiaohan Yan}\textsuperscript{2,\textdagger} \quad
\textbf{Zunian Wan}\textsuperscript{2} \quad
\textbf{Xuan Du}\textsuperscript{2} \\
\textbf{Zhaozhan Chi}\textsuperscript{2} \quad
\textbf{Jianbo Mao}\textsuperscript{2} \quad
\textbf{Ruipu Wu}\textsuperscript{2} \quad
\textbf{Rushuai Yang}\textsuperscript{2} \\
\textbf{Houlin Li}\textsuperscript{2} \quad
\textbf{Shukai Yang}\textsuperscript{2} \quad
\textbf{Jing Wu}\textsuperscript{2} \quad
\textbf{Yuxiang Yan}\textsuperscript{2} \\
\textbf{Yongcheng Liu}\textsuperscript{2} \quad
\textbf{Chuankang Li}\textsuperscript{2} \quad
\textbf{Guanghui Ren}\textsuperscript{2} \quad
\textbf{Wei Shan}\textsuperscript{2} \quad
\textbf{Maoqing Yao}\textsuperscript{2}
\\
\textsuperscript{1}South China University of Technology \quad
\textsuperscript{2}AgiBot \quad
\\
\textsuperscript{\textdagger}Corresponding Author
}

\begin{document}

\pagestyle{empty}

\noindent
\twocolumn[{
\renewcommand\twocolumn[1][]{#1}

\maketitle

\vspace{-4mm}
\begin{center}
    \centering
    \captionsetup{type=figure}
    \includegraphics[width=\textwidth]{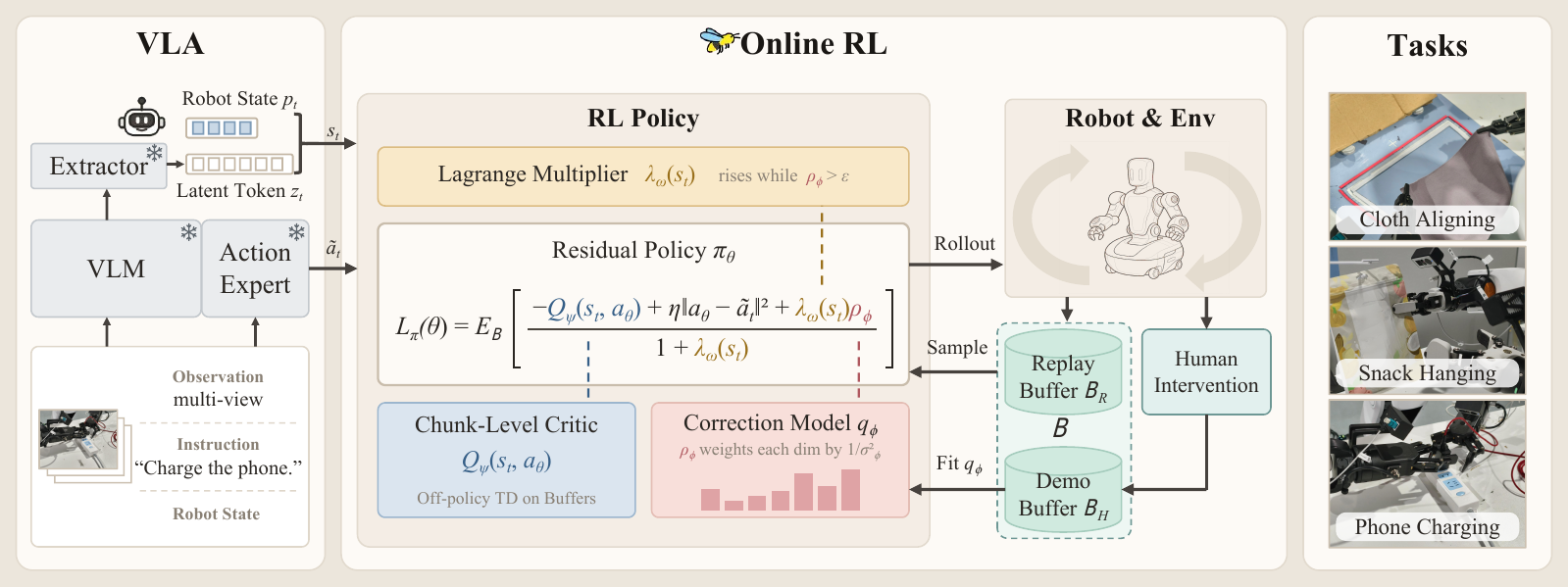}
    \captionof{figure}{\textbf{Overview of \ours{}.}
    \ours{} converts human corrections of VLA proposals into an uncertainty-aware
    constraint on policy optimization.
    \textit{Left:} a frozen VLA produces an action proposal
    $\tilde{\mathbf{a}}_t$ while an extractor maps the VLA's internal
    features to a latent token that, with the robot state $p_t$,
    forms the RL state $\mathbf{s}_t$.
    \textit{Middle:} rather than imitating interventions, \ours{} distills
    them via a Correction Model $q_\phi$ into an adaptive constraint
    $\rho_\phi$ that weights each action dimension by the inverse of its
    predicted correction variance $\sigma^2_\phi$. The Residual Policy
    $\pi_\theta$ refines $\tilde{\mathbf{a}}_t$ into $\mathbf{a}_\theta$ by
    optimizing $L_\pi(\theta)$, which balances $Q_\psi$ and a VLA anchor ($\eta$) against $\rho_\phi$
    weighted by a Lagrange multiplier $\lambda_\omega(\mathbf{s}_t)$.
    \textit{Right:} we evaluate \ours{} on three real-world tasks.}
    \label{fig:main}
\end{center}

}]

\begin{abstract}
Vision-language-action (VLA) models handle long-horizon manipulation, yet success hinges on a few precision-critical phases where millimeter-scale errors undo all prior progress. Online reinforcement learning (RL) can optimize exactly these actions, but free exploration is far too costly on real robots, which makes human corrections indispensable. However, existing online RL methods for VLAs either cannot incorporate such corrections or fold them into undifferentiated supervision. Yet human corrections are not uniformly noisy but reliable along some action dimensions and variable along others. Building on this, we introduce \ours{}, an intervention-adaptive framework for real-world RL on a frozen VLA that lets the policy go \textbf{BE}yond \textbf{E}xpert imitation. We formulate human corrections not as actions to reproduce but as evidence about a constraint: a Correction Model predicts how a human would correct a given VLA proposal and how consistent the correction is along each action dimension. This predicted consistency sets the per-dimension tightness of a constraint on policy optimization. Where corrections are consistent the policy stays close to the human, and where they vary, the constraint relaxes. We evaluate \ours{} on three real-world manipulation tasks and one LIBERO-Pro simulation task at a matched online-data budget. \ours{} attains the highest success rate on every task, 91.2\% on average against 57.5\% for RLT and 42.1\% for DSRL, and the lowest human intervention rate on all real-world tasks.
\end{abstract}

\section{Introduction}
\label{sec:intro}

Vision-language-action (VLA) models have emerged as capable generalist controllers for robotic manipulation, leveraging broad behavioral priors and semantic representations to execute long-horizon, multi-stage tasks from visual observations and language instructions~\cite{rt2,openvla,pi0,gr00t,pi05,pi06}.
Yet task success often hinges on a few precision-critical phases such as alignment and insertion, where millimeter-scale pose errors invalidate all preceding progress.
This brittleness makes online reinforcement learning (RL) attractive. By interacting with the target environment, RL can directly optimize the actions that determine task success~\cite{hilserl,rl100}.
On real robots, however, every rollout costs time and hardware life, and sparse rewards coupled with high-dimensional action chunks limit unconstrained exploration, making targeted human corrections a valuable but scarce resource.

However, existing online RL methods for VLAs either provide no mechanism for human corrections or use them without distinction.
DSRL~\cite{dsrl} steers the frozen VLA by searching its latent noise space and has no interface for human input at all.
The recent RL Token (RLT) framework~\cite{rlt} stores human interventions and autonomous rollouts in a shared replay buffer. When the human intervenes, RLT replaces the VLA proposal stored for that transition with the human correction, so its actor is trained with human corrections in place of VLA proposals on intervention samples, whereas its proposal input always comes from the VLA at deployment.
Prior work such as SiLRI~\cite{silri} adjusts the overall imitation strength with a state-wise uncertainty estimate but still treats all action dimensions uniformly.
Human corrections, however, may be consistent in some action dimensions yet variable in others, motivating dimension-wise rather than uniform constraints.

To this end, we introduce \ours{}, an intervention-adaptive framework for VLA-anchored real-world RL that retains the VLA proposal associated with each human correction and models the correction as a residual from that proposal. Based on the learned residual distribution, \ours{} imposes a dimension-wise, uncertainty-aware constraint on policy optimization, allowing the policy to go \textbf{BE}yond \textbf{E}xpert imitation.
As illustrated in Figure~\ref{fig:main}, \ours{} draws two outputs from a single frozen VLA.
An extractor converts the features of the VLA into a latent token that supplies task-relevant input to online RL.
The action expert produces an action chunk that is retained as a deployment-time proposal, which a lightweight chunk-level \emph{Residual Policy} then refines into the action executed by the robot.
Rather than imitating human interventions directly, \ours{} uses them to train a separate \emph{Correction Model}.
Conditioned on the current state and on the VLA proposal, the Correction Model predicts a distribution over the correction residual, from which a mean and an uncertainty are obtained for each action dimension.
Adding the predicted mean to the VLA proposal yields the corrected action, which serves as the center of a correction constraint on the policy, whereas the predicted uncertainty measures how consistent human corrections are in each action dimension.
RL then optimizes the Residual Policy subject to this constraint, which keeps the policy close to the corrected action along consistent dimensions while allowing the constraint to relax along variable ones.
Overall, our contributions are as follows:
\begin{itemize}
    \item We introduce \ours{}, an intervention-adaptive framework for VLA-anchored real-world RL that converts human interventions into a proposal-conditioned correction prior while preserving the long-horizon competence of the frozen VLA.

    \item We develop a dimension-wise, uncertainty-aware policy constraint that adapts to the consistency of human corrections in each action dimension.

    \item We evaluate \ours{} on three real-world manipulation tasks and on a LIBERO-Pro simulation task. \ours{} attains the highest success rate on every task and the lowest human intervention rate on all real-world tasks.

\end{itemize}

\section{Related Work}
\label{sec:related}

\subsection{RL for Vision-Language-Action Models}
\label{sec:related_vla_rl}

Reinforcement learning increasingly extends pretrained VLA policies beyond their demonstration data, through large-scale simulated rollouts~\cite{vlarl,riptvla,flare}, preference alignment~\cite{grape}, or distillation of task-specific RL controllers~\cite{rldg}. On real hardware, ConRFT~\cite{conrft} trains a consistency-based action head offline and then online, and iRe-VLA~\cite{irevla} alternates supervised and actor--critic updates, yet both fine-tune the VLA rather than keep its output as a fixed proposal for online correction. DSRL~\cite{dsrl} instead freezes the base policy and steers it toward higher return by searching its latent noise space.

RLT~\cite{rlt} freezes the VLA and trains a chunk-level actor--critic that takes a latent token together with the explicit action proposal of the VLA as input. Because the learned policy is kept close to this proposal, the resulting update amounts to value maximization regularized toward a behavior prior~\cite{awac,td3bc}, with the frozen VLA serving as that prior.
\ours{} builds on this interface with a residual policy~\cite{respolicy,resrl} but focuses on how human corrections should guide policy improvement.

\subsection{Human-in-the-Loop RL}
\label{sec:related_hil_rl}
Human-in-the-loop RL improves the exploration efficiency of real-world RL by allowing a human operator to guide the policy during online interaction. These methods differ mainly in when the human takes over and in how the resulting data are reused. DAgger~\cite{dagger} relabels on-policy states with actions queried from an always-available human, whereas HG-DAgger~\cite{hgdagger} and ThriftyDAgger~\cite{thriftydagger} gate takeovers by human judgment or by a risk budget. Intervention-weighted schemes instead upweight corrective segments~\cite{iwr,sirius}, and other formulations treat the intervention signal as reward or value information rather than as an imitation target~\cite{rlif,pvp}. Building on off-policy learning from mixed offline and online data~\cite{rlpd}, SERL~\cite{serl} and RL-100~\cite{rl100} provide practical training pipelines, and HIL-SERL~\cite{hilserl} further incorporates demonstrations and human takeovers at failure-prone states, which supplies corrective transitions that autonomous exploration rarely encounters. Such targeted assistance is valuable when robot time is scarce, yet these pipelines train task-specific visuomotor networks~\cite{act,diffusionpolicy} and therefore leave the competence of a pretrained VLA unused.

SiLRI~\cite{silri} accounts for noisy human interventions through a state-conditioned human behavior model and a state-wise Lagrange multiplier that balances return maximization against imitation. The constraint of SiLRI, however, is a single scalar bound per state, so it cannot express how the reliability of a human correction varies from one action dimension to the next. In contrast, \ours{} predicts a per-dimension correction distribution in the spirit of heteroscedastic uncertainty estimation~\cite{nixweigend,kendallgal}, so that the resulting Mahalanobis constraint adapts across action dimensions, while the overall constraint strength is still set by a state-dependent multiplier trained by dual ascent~\cite{rcpo}.

\section{Preliminaries}
\label{sec:prelim}

\subsection{Chunk-Level RL over a VLA Interface}
\label{subsec:prelim_rl_interface}

We formulate task execution as a chunk-level RL problem. At each chunk boundary $t$, the agent observes a state $\mathbf{s}_t$ and emits an action chunk $\mathbf{a}_t\in\mathbb{R}^{D}$, which concatenates $K$ primitive actions each of dimension $d$ ($D{=}Kd$). The environment returns a reward $r_t$, and we maximize the expected discounted return $J(\pi)=\mathbb{E}_{\pi}\big[\sum_{t}\gamma^{t} r_t\big]$ with discount factor $\gamma\in(0,1)$.

Following RLT~\cite{rlt}, we ground this state and action space in a frozen VLA while retaining its action proposal. At each boundary $t$, the frozen VLA $\pi_{\mathrm{vla}}$ maps the current observation $\mathbf{o}_t$, language instruction $g$, and proprioceptive state $\mathbf{p}_t$ to an action proposal $\tilde{\mathbf{a}}_t \sim \pi_{\mathrm{vla}}(\cdot\mid \mathbf{o}_t,g,\mathbf{p}_t)$. An Extractor converts the VLA's internal features into a latent token $\mathbf{z}_t$, giving the RL state $\mathbf{s}_t = [\,\mathbf{z}_t \,\Vert\, \mathbf{p}_t\,]$. The pair $(\mathbf{s}_t,\tilde{\mathbf{a}}_t)$ is the fixed VLA--RL interface, with the critic evaluating chunks at $\mathbf{s}_t$ and the policy additionally conditioning on $\tilde{\mathbf{a}}_t$.

A chunk-level critic $Q_\psi(\mathbf{s}_t,\mathbf{a}_t)$ evaluates an entire chunk rather than a single step~\cite{rlt}, trained off-policy on transitions sampled from a buffer $\mathcal{B}$ with the $K$-step Bellman target
\begin{equation}
    y_t = R_t + \gamma^{K}\,(1-\delta_t)\, Q_{\bar\psi}(\mathbf{s}_{t+K},\mathbf{a}_{t+K}),
    \label{eq:kstep_target}
\end{equation}
where $R_t=\sum_{j=0}^{K-1}\gamma^{j} r_{t+j}$ is the discounted chunk return, $\delta_t$ indicates termination within the chunk, $\mathbf{a}_{t+K}$ is the next chunk produced by the current policy, and $Q_{\bar\psi}$ is a target critic. The critic minimizes the Bellman error
\begin{equation}
    L_Q(\psi)=\mathbb{E}_{\mathcal{B}}\big[(Q_\psi(\mathbf{s}_t,\mathbf{a}_t)-y_t)^2\big].
    \label{eq:critic_loss}
\end{equation}

\subsection{Learning from Human Interventions}
\label{subsec:prelim_human_interventions}

In human-in-the-loop RL~\cite{hilserl}, a human correction $\mathbf{a}^H_t$ overrides the policy action and becomes the executed action $\mathbf{a}_t$. All online transitions, autonomous or human-corrected, go to the replay buffer $\mathcal{B}_R$, and human-corrected chunks are additionally copied into a demonstration buffer $\mathcal{B}_H$. Each training batch is drawn in equal parts from the two, and $\mathcal{B}$ denotes this training distribution. The VLA and the encoder stay frozen throughout, and rollout collection, human interventions, and off-policy updates interleave at each chunk boundary. \ours{} builds on this off-policy critic and its two buffers, turning human interventions into an uncertainty-aware constraint on the policy.

\section{Method}
\label{sec:method}

\subsection{Problem Formulation}
\label{subsec:problem_formulation}

At each chunk boundary, the frozen VLA--RL interface provides a state--proposal
pair $(\mathbf{s}_t,\tilde{\mathbf{a}}_t)$, where
$\mathbf{s}_t=[\mathbf{z}_t\mathbin\Vert\mathbf{p}_t]$. The proposal is an
additional input to the online learner. We do not propagate gradients through
the frozen VLA.

A deterministic Residual Policy predicts an action residual and produces
$\mathbf{a}_\theta=\tilde{\mathbf{a}}_t+
\boldsymbol\Delta^\pi_\theta(\mathbf{s}_t,\tilde{\mathbf{a}}_t)$. Meanwhile, a
proposal-conditioned Gaussian Correction Model $q_\phi$ represents how a human
would correct the same proposal. Let $\boldsymbol\mu_\phi$ and
$\boldsymbol\Sigma_{\phi,t}=\operatorname{diag}(\boldsymbol\sigma_\phi^2)$
denote its predicted mean residual and diagonal covariance, respectively. They
define the predicted corrected action
$\hat{\mathbf{a}}^H_t=\tilde{\mathbf{a}}_t+\boldsymbol\mu_\phi$ and the
normalized Mahalanobis deviation
\begin{equation}
    \rho_\phi
    =\frac{1}{D}
    (\mathbf{a}_\theta-\hat{\mathbf{a}}^H_t)^\top
    \boldsymbol\Sigma_{\phi,t}^{-1}
    (\mathbf{a}_\theta-\hat{\mathbf{a}}^H_t).
    \label{eq:mahalanobis_constraint}
\end{equation}
Thus, consistent human corrections define narrow feasible directions, whereas
variable corrections leave more freedom for return-driven improvement.

Our objective is to improve task return without allowing the learned policy to
depart arbitrarily from this correction-informed feasible set. We formulate
policy learning as
\begin{equation}
\begin{aligned}
    \min_{\theta}\quad
    &\mathbb{E}_{\mathcal B}\!\left[
      -Q_\psi(\mathbf{s}_t,\mathbf{a}_\theta)
      +\eta\lVert\mathbf{a}_\theta-\tilde{\mathbf{a}}_t\rVert_2^2
    \right] \\
    \mathrm{s.t.}\quad
    &\rho_\phi\le\varepsilon,
\end{aligned}
    \label{eq:constrained_policy_problem}
\end{equation}
where the constraint is evaluated on state--proposal pairs sampled from
$\mathcal B$, and $\varepsilon>0$ is fixed. The predicted correction
mean centers the feasible set, and its variances determine the set's
dimension-wise geometry. Since $q_\phi$ is Gaussian, for a fixed Correction
Model, $\rho_\phi$ is proportional to the actor-dependent term of
$-\log q_\phi(\mathbf{a}_\theta-\tilde{\mathbf{a}}_t\mid
\mathbf{s}_t,\tilde{\mathbf{a}}_t)$.

\subsection{Learning the Correction Distribution}
\label{subsec:correction_distribution}

Each human-corrected sample copied into the demonstration buffer
$\mathcal{B}_H$ retains both the VLA proposal $\tilde{\mathbf{a}}_t$ and the
human correction $\mathbf{a}^H_t$, rather than replacing the proposal with the
correction. With correction residual
$\boldsymbol\Delta^H_t=\mathbf{a}^H_t-\tilde{\mathbf{a}}_t$, the Correction
Model predicts
\begin{equation}
    q_\phi(\boldsymbol\Delta^H_t\mid
    \mathbf{s}_t,\tilde{\mathbf{a}}_t)
    =\mathcal{N}\!\left(
        \boldsymbol\mu_\phi(\mathbf{s}_t,\tilde{\mathbf{a}}_t),
        \boldsymbol\Sigma_{\phi,t}
    \right).
    \label{eq:correction_model}
\end{equation}
It is trained only on $\mathcal{B}_H$ by negative log-likelihood,
\begin{equation}
    \mathcal{L}_{\mathrm{corr}}(\phi)
    =-\mathbb{E}_{\mathcal{B}_H}
    \!\left[\log q_\phi(\boldsymbol\Delta^H_t\mid
    \mathbf{s}_t,\tilde{\mathbf{a}}_t)\right].
    \label{eq:correction_nll}
\end{equation}
The proposal is kept as an explicit model input so that the Correction Model
need not recover that high-dimensional action from the compact RL state. During data collection, small Gaussian exploration
noise is added to $\mathbf{a}_\theta$ as in TD3~\cite{td3}.

\subsection{State-Dependent Primal--Dual Optimization}
\label{subsec:primal_dual_optimization}

Introducing a nonnegative state-dependent multiplier
$\lambda(\cdot)\geq 0$ for
Eq.~\eqref{eq:constrained_policy_problem} yields the population Lagrangian
\begin{equation}
    \begin{aligned}
    \mathcal{L}(\theta,\lambda)
    ={}&\mathbb{E}_{\mathcal B}\!\left[
      -Q_\psi(\mathbf{s}_t,\mathbf{a}_\theta)\right.\\
      &\left.\quad+\eta\lVert\mathbf{a}_\theta-\tilde{\mathbf{a}}_t\rVert_2^2
      +\lambda(\mathbf{s}_t)(\rho_\phi-\varepsilon)
    \right].
    \end{aligned}
    \label{eq:policy_lagrangian}
\end{equation}
with saddle problem $\min_\theta\max_{\lambda(\cdot)\ge0}
\mathcal{L}(\theta,\lambda)$. We approximate $\lambda$ by
$\lambda_\omega(\mathbf{s}_t)$ with a softplus output. For the actor step,
$-\lambda_\omega\varepsilon$ is constant in $\theta$ and can be omitted. We
further normalize each sampled state by $1+\lambda_\omega(\mathbf{s}_t)$ ~\cite{silri}, giving
\begin{equation}
    \mathcal{L}_{\pi}(\theta)
    =\mathbb{E}_{\mathcal{B}}\!\left[
        \frac{
            -Q_\psi(\mathbf{s}_t,\mathbf{a}_\theta)
            +\eta\lVert\mathbf{a}_\theta-\tilde{\mathbf{a}}_t\rVert_2^2
            +\lambda_\omega(\mathbf{s}_t)\rho_\phi
        }{
            1+\lambda_\omega(\mathbf{s}_t)
        }
    \right].
    \label{eq:policy_objective}
\end{equation}
The normalization stabilizes large-multiplier updates.

The multiplier is trained by dual ascent, implemented as minimization of
\begin{equation}
    \mathcal{L}_{\lambda}(\omega)
    =-\mathbb{E}_{\mathcal{B}}\!\left[
        \lambda_\omega(\mathbf{s}_t)(\rho_\phi-\varepsilon)
    \right].
    \label{eq:multiplier_loss}
\end{equation}
Human corrections therefore shape how the policy is constrained rather than
serving as direct policy targets.

\subsection{Training Procedure}
\label{subsec:training_procedure}

The Correction Model is pre-trained on the corrections that seed
$\mathcal{B}_H$ and then updated only after every $M$ new intervention samples,
for $NM$ gradient steps at a fixed update-to-data ratio $N$. This conservative
schedule reduces variance collapse on the small intervention buffer; full
details appear in Section~\ref{sec:supp_impl}. Algorithm~\ref{alg:bee}
summarizes the complete online training loop.

\begin{algorithm}[t]
\caption{\ours{}: Intervention-Adaptive Policy Optimization}
\label{alg:bee}
\begin{algorithmic}[1]
\REQUIRE Frozen VLA $\pi_{\mathrm{vla}}$; constraint bound $\varepsilon$, anchor weight $\eta$, target rate $\tau$
\STATE Initialize $\pi_\theta$, $q_\phi$, $\lambda_\omega$, $Q_\psi$, and target critic $\bar\psi\leftarrow\psi$; seed $\mathcal{B}_H$ with pre-collected correction transitions
\STATE Pre-train $q_\phi$ on $\mathcal{B}_H$ by Eq.~\eqref{eq:correction_nll}
\FOR{each chunk boundary $t$}
    \STATE Obtain $(\mathbf{s}_t,\tilde{\mathbf{a}}_t)$ and compute $\mathbf{a}_\theta=\tilde{\mathbf{a}}_t+\boldsymbol\Delta^\pi_\theta(\mathbf{s}_t,\tilde{\mathbf{a}}_t)$
    \IF{human intervenes}
        \STATE Execute $\mathbf{a}^H_t$
    \ELSE
        \STATE Execute $\mathbf{a}_\theta$ with exploration noise
    \ENDIF
    \STATE Store the full transition in $\mathcal{B}_R$ and, if corrected, copy it to $\mathcal{B}_H$ while retaining $(\tilde{\mathbf{a}}_t,\mathbf{a}^H_t)$
    \STATE Sample a minibatch $\mathcal{D}$ in equal parts from $\mathcal{B}_R$ and $\mathcal{B}_H$
    \STATE Update $Q_\psi$ on $\mathcal{D}$ by Eq.~\eqref{eq:critic_loss}; soft-update $\bar\psi\leftarrow(1-\tau)\bar\psi+\tau\psi$
    \STATE Every $M$ new corrections, update $q_\phi$ by $NM$ steps on Eq.~\eqref{eq:correction_nll}
    \STATE On $\mathcal{D}$, form $\hat{\mathbf{a}}^H$ and evaluate $\rho_\phi$ by Eq.~\eqref{eq:mahalanobis_constraint}
    \STATE Update $\pi_\theta$ by Eq.~\eqref{eq:policy_objective} and $\lambda_\omega$ by Eq.~\eqref{eq:multiplier_loss} on $\mathcal{D}$
\ENDFOR
\end{algorithmic}
\end{algorithm}

\section{Experiments}
\label{sec:experiments}

\begin{table*}[t]
\centering
\caption{\textbf{Comparison of \ours{} with the Base Policy, RLT, and DSRL}. Success rate (SR, \%, $\uparrow$) and human intervention rate (Int., \%, $\downarrow$) on the real-world tasks phone charging, snack hanging, and cloth aligning, and on the LIBERO-PRO simulation task bowl placing. The Base Policy and DSRL involve no online intervention, so no intervention rate is reported.}
\label{tab:main_results}
\resizebox{\textwidth}{!}{%
\begin{tabular}{l cc cc cc cc cc}
\toprule
\multirow{2}{*}{\textbf{Method}}
& \multicolumn{2}{c}{\textbf{Phone Charging}}
& \multicolumn{2}{c}{\textbf{Snack Hanging}}
& \multicolumn{2}{c}{\textbf{Cloth Aligning}}
& \multicolumn{2}{c}{\textbf{Bowl Placing}}
& \multicolumn{2}{c}{\textbf{Overall}} \\
\cmidrule(lr){2-3}\cmidrule(lr){4-5}\cmidrule(lr){6-7}\cmidrule(lr){8-9}\cmidrule(lr){10-11}
& SR\,$\uparrow$ & Int.\,$\downarrow$
& SR\,$\uparrow$ & Int.\,$\downarrow$
& SR\,$\uparrow$ & Int.\,$\downarrow$
& SR\,$\uparrow$ & Int.\,$\downarrow$
& SR\,$\uparrow$ & Int.\,$\downarrow$ \\
\midrule
Base Policy      & 90.0 $\pm$ 5.0 & -- & 0.0 $\pm$ 0.0 & -- & 36.7 $\pm$ 7.6 & -- & 28.3 $\pm$ 7.6 & -- & 38.8 & -- \\
RLT              & 58.3 $\pm$ 2.9$^{\dagger}$ & 22.5 & 21.7 $\pm$ 5.8 & 62.1 & 73.3 $\pm$ 7.6 & 85.3 & 76.7 $\pm$ 17.6 & \textbf{26.2} & 57.5 & 49.0 \\
DSRL             & 93.3 $\pm$ 2.9 & -- & 13.3 $\pm$ 2.9 & -- & 31.7 $\pm$ 2.9 & -- & 30.0 $\pm$ 5.0 & -- & 42.1 & -- \\
\rowcolor{gray!12}
\textbf{\ours{}} & \textbf{100.0 $\pm$ 0.0} & \textbf{12.2} & \textbf{85.0 $\pm$ 0.0} & \textbf{17.0} & \textbf{90.0 $\pm$ 10.0} & \textbf{65.3} & \textbf{90.0 $\pm$ 13.2} & 27.4 & \textbf{91.2} & \textbf{30.5} \\
\bottomrule
\end{tabular}%
}
\\[2pt]
{\footnotesize\raggedright $^{\dagger}$Reported at the matched online-data budget used for all methods. RLT requires roughly $3\times$ the online data to reach $96.7 \pm 5.8$.\par}
\end{table*}

\subsection{Experimental Setup}
\label{sec:exp_setup}

\begin{figure}[t]
\centering
\includegraphics[width=\columnwidth]{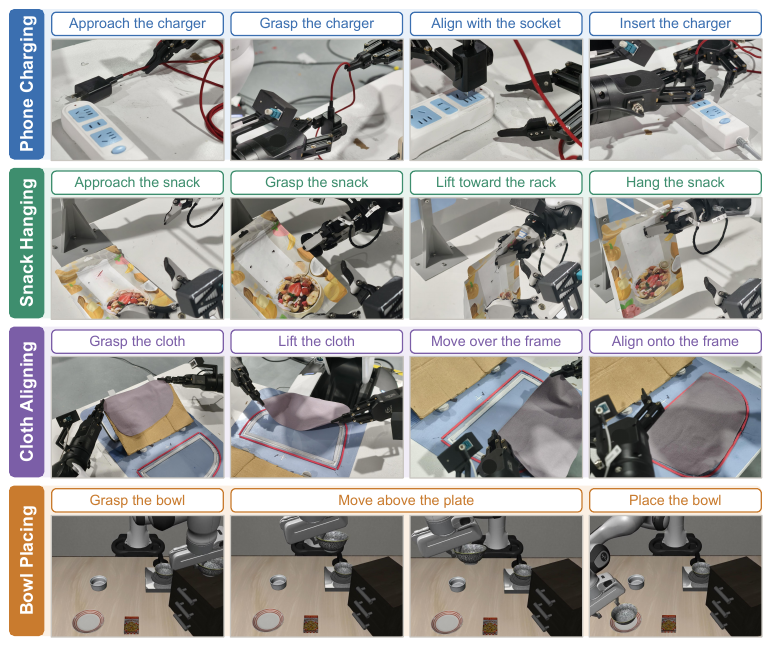}
\caption{\textbf{Evaluation Tasks.}
Keyframes of the three real-world tasks (phone charging, snack hanging, cloth aligning) and of the LIBERO-PRO simulation task, bowl placing.}
\label{fig:tasks}
\end{figure}

\paragraph{Real-world tasks}
We evaluate three real-world manipulation tasks, as demonstrated in Figure~\ref{fig:tasks}, each
ending in a precision-critical contact phase in which small action errors
determine success:
\begin{itemize}
    \item \textbf{Phone charging} The left arm grasps and lifts the cable while
    the right arm approaches and grasps the charger, aligns it with the socket,
    and inserts it, which demands millimeter-level alignment under partial
    observability of the socket and the charger tip.
    \item \textbf{Snack hanging.} The robot approaches and grasps a deformable
    snack package, lifts it toward the rack, and hangs it by the hole in the
    package. The package deforms under the grasp, and the resulting pose shifts
    the target location across episodes.
    \item \textbf{Cloth aligning.} The robot grasps a piece of cloth, lifts it,
    moves it over the target frame, and aligns it onto that frame. Deformation
    during transport makes the final contact and release the decisive step.
\end{itemize}
Each task spans $30$--$60$\,s of execution, i.e., roughly $900$--$1800$ control
steps at $30$\,Hz.

\paragraph{Simulation task}
We additionally evaluate \textbf{bowl placing} in LIBERO-PRO~\cite{liberopro},
an evaluation extension of the LIBERO benchmark~\cite{libero}, in which the arm
grasps a bowl, moves it above the plate, and places it down over a comparable
horizon. The benchmark configuration is detailed in
Appendix~\ref{sec:supp_tasks}.

\paragraph{Implementation details}
We fine-tune the open-source $\pi_{0.5}$ weights with behavior cloning on each task, then freeze the resulting VLA together with an RLT-style~\cite{rlt} latent-token encoder as the fixed VLA--RL interface. During online RL, only the lightweight Residual Policy, the critic, the Correction Model, and the state-dependent multiplier are trained, all off-policy at the chunk level under an action normalization shared across methods. The latter three are training-time components: at evaluation and deployment, interventions and exploration noise are switched off and the robot is driven by the Residual Policy on top of the frozen VLA interface, at no additional inference cost. All four tasks are long-horizon. For phone charging and cloth aligning, which span the longest horizons, online RL refines only their precision-critical phase, whereas the relatively shorter snack hanging and bowl placing are trained over the full task. The real-world robot is controlled at 30\,Hz with a task-dependent action space: on phone charging, the per-timestep action is 16-dimensional (14 dual-arm joint dimensions, 7 per arm, plus 2 end-effector dimensions), and a chunk length of $K{=}10$ .  All tasks use a sparse task-completion reward, except cloth aligning with a vision-based dense reward. Further task, reward, and implementation details appear in Appendices~\ref{sec:supp_tasks}--\ref{sec:supp_impl}.

\paragraph{Baselines}
We compare \ours{} against three baselines that operate in the same frozen-VLA setting. The Base Policy is the $\pi_{0.5}$~\cite{pi05} VLA after supervised fine-tuning (SFT), evaluated before any online adaptation as a pre-adaptation reference. RLT~\cite{rlt} performs chunk-level online RL over the frozen VLA's latent token, regularized toward the VLA proposal. DSRL~\cite{dsrl} keeps the VLA fixed and steers it by searching its latent noise space. We additionally evaluate DAgger~\cite{dagger}, SiLRI~\cite{silri}, and HIL-SERL~\cite{hilserl} on a subset of our tasks and report them in Table~\ref{tab:supp_more_baselines} of Appendix~\ref{sec:supp_baselines}.

\paragraph{Protocol and metrics}
All methods start from the same fine-tuned VLA and are compared at a matched robot-data budget of around 20 pre-collected episodes of human corrections plus a fixed number of online episodes per task. RLT and \ours{}, the two methods that train with human intervention, are seeded with those 20 episodes and follow the same intervention protocol, whereas DSRL maintains no demonstration buffer and receives them as additional online interaction. The success rate is measured in three evaluation rounds of 20 trials per task and reported as the mean and standard deviation across rounds. Phone charging and cloth aligning report success only on their precision-critical phase, whereas snack hanging and bowl placing report full-task success. The human intervention rate is the proportion of control steps taken over by the human during online training. Full budget accounting and intervention details are provided in Appendix~\ref{sec:supp_protocol}.

\begin{figure}[t]
\centering
\resizebox{0.85\columnwidth}{!}{%
  \includegraphics{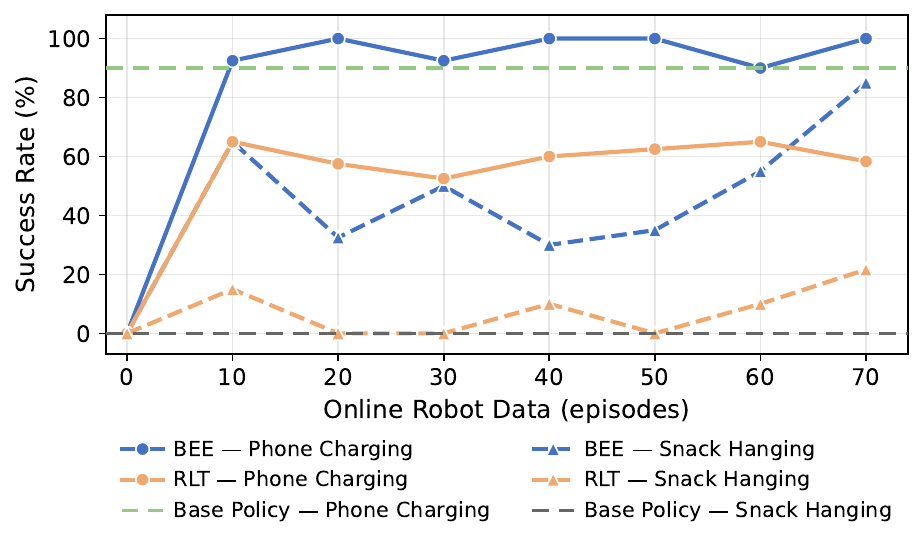}%
}
\caption{\textbf{Sample Efficiency on Phone Charging and Snack Hanging}. Success rate against the number of online robot episodes, with color distinguishing \ours{} from RLT and line style distinguishing the two tasks; the horizontal dashed lines mark the Base Policy. The right end of the axis is the matched online-data budget at which Table~\ref{tab:main_results} is reported. \ours{} stays above RLT at every budget on both tasks.}
\label{fig:sample_efficiency}
\end{figure}

\begin{figure}[t]
\centering
\includegraphics[width=\columnwidth]{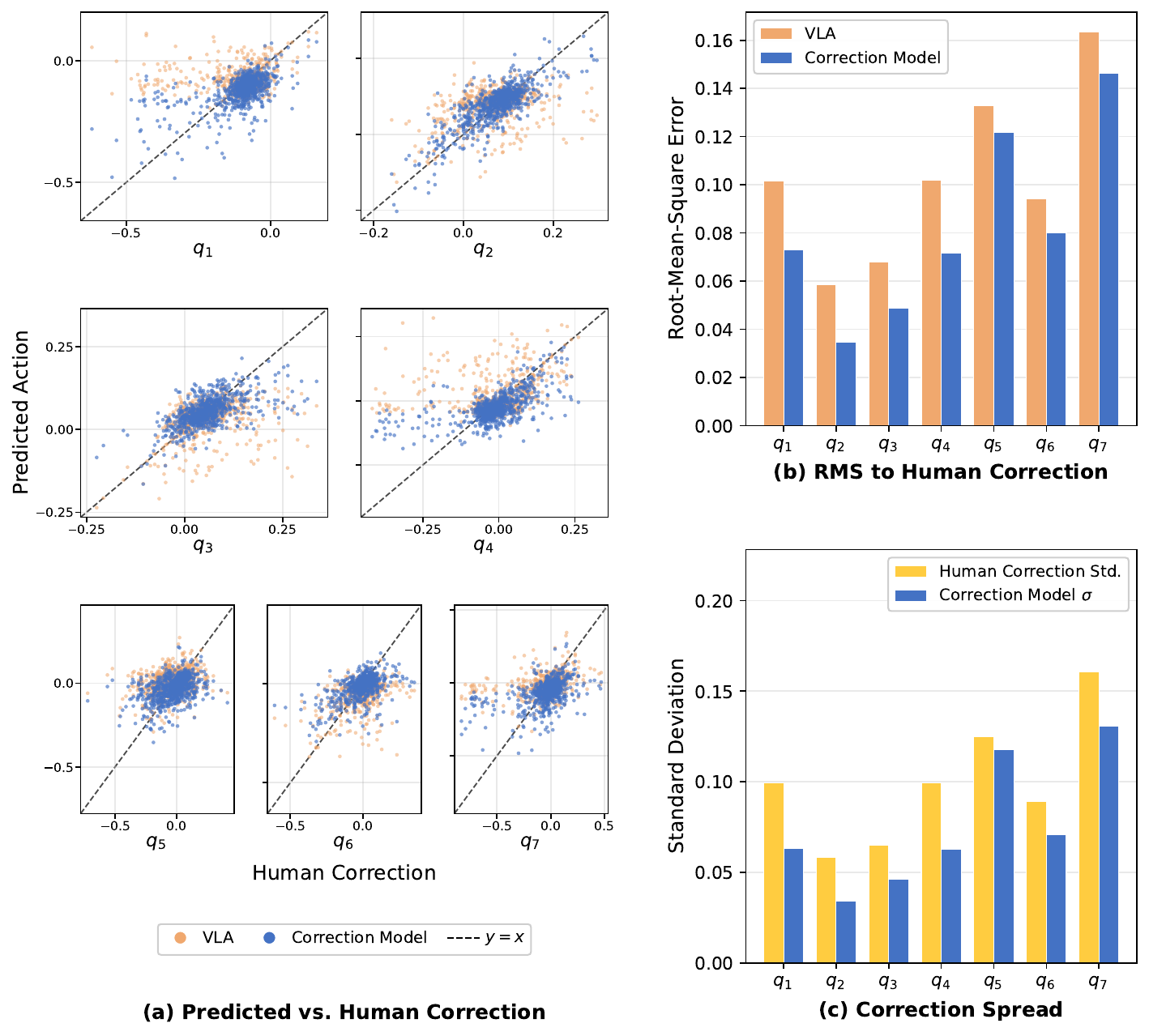}
\caption{\textbf{Correction Model Calibration on Phone Charging}, evaluated on human-intervention demonstrations over the seven joint dimensions $q_1$--$q_7$ of the intervened arm. \textbf{(a)}~The VLA proposal $\tilde{\mathbf{a}}_t$ and the corrected action $\hat{\mathbf{a}}^H_t$ produced by the Correction Model, plotted against the human correction. Points closer to the $y{=}x$ line match it better. \textbf{(b)}~Per-dimension root-mean-square error to the human correction. The corrected action reduces this error below the VLA proposal in every dimension. \textbf{(c)}~Predicted per-dimension standard deviation of the Correction Model and empirical standard deviation of human corrections. The prediction follows the same per-dimension pattern as the empirical spread.}
\label{fig:correction_model}
\end{figure}

\begin{figure}[t]
\centering
\includegraphics[width=\columnwidth]{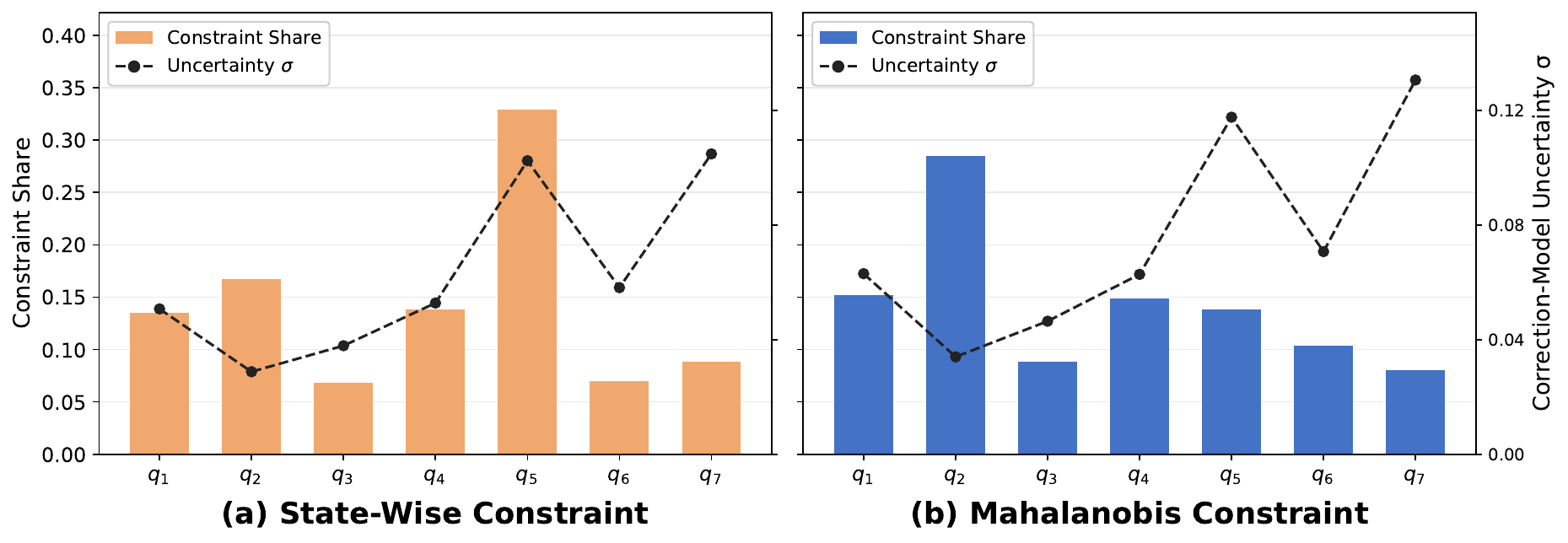}
\caption{\textbf{Dimension-wise Constraint Allocation on Phone Charging}. Bars show each dimension's constraint share (its normalized contribution to $\rho_\phi$), and the overlaid curve is the predicted uncertainty $\sigma$ of the Correction Model. \textbf{(a)}~The state-wise form weights all dimensions equally, so a high-uncertainty dimension such as $q_5$ claims the largest share. \textbf{(b)}~The Mahalanobis form concentrates the share on low-uncertainty, consistently corrected dimensions (e.g., $q_2$) and downweights high-uncertainty ones (e.g., $q_5$).}
\label{fig:constraint_alloc}
\end{figure}

\subsection{Results and Analysis}
\label{sec:exp_results}

\paragraph{Overall comparison}
Table~\ref{tab:main_results} compares \ours{} with the Base Policy, RLT, and DSRL across phone charging, snack hanging, cloth aligning, and bowl placing. \ours{} attains the highest success rate on every task and the lowest intervention rate on all three real-world tasks as well as overall. The gains hold across very different Base-Policy starting points. Where the Base Policy is already strong, \ours{} improves it further to a saturated success rate within the matched online-data budget, and where it starts near zero, \ours{} still reaches a comparably high success rate. The other methods improve far less under the same budget, in some cases even falling below the Base Policy, and RLT does so at a substantially higher intervention rate. Additional comparisons with DAgger, SiLRI, and HIL-SERL are reported in Table~\ref{tab:supp_more_baselines} of Appendix~\ref{sec:supp_baselines}; none reaches competitive performance under the same budget.

\paragraph{Sample efficiency}
Figure~\ref{fig:sample_efficiency} tracks the success rate over online episodes on phone charging and snack hanging. On phone charging, where the Base Policy is already strong, \ours{} matches it within the first few episodes and then saturates. On snack hanging, where the Base Policy never succeeds, \ours{} first explores away from it, which temporarily lowers the success rate before it converges to a high one well inside the same budget. RLT improves early on phone charging but then plateaus below the Base Policy, and on snack hanging it stays close to the Base Policy throughout. The two methods also differ in how the robot behaves while learning. Because \ours{} anchors every update to the VLA proposal and tightens the correction constraint only along the dimensions humans correct consistently, its motion stays smooth and human-like from the first episodes onward, in training as in evaluation, whereas RLT visibly jitters early in training before settling. Smoother early behavior in turn calls for fewer human corrections, and \ours{} indeed attains the lowest intervention rate on all three real-world tasks, as reported in Table~\ref{tab:main_results}. Taken together, these differences reflect how efficiently each method turns online robot data and human corrections into usable behavior.

\paragraph{Per-dimension reliability of human corrections}
Dimension-wise weighting assumes that human corrections are reliable on some action dimensions but noisy on others. We test this on held-out human-intervention demonstrations, restricted to the seven joint dimensions of the intervened arm ($q_1$--$q_7$), because the other arm and the end-effector receive essentially no human correction. As Figure~\ref{fig:correction_model}c shows, the empirical spread of the corrections indeed varies by about $3\times$ across dimensions, from the most consistently edited ($q_2$) to the most variable ($q_7$). The corrections are nonetheless predictable from the state and the VLA proposal. Recentered by the Correction Model, the corrected action lies closer to the $y{=}x$ line than the VLA proposal and reduces the per-dimension error to the human correction in every dimension, as Figure~\ref{fig:correction_model}a,b illustrates. Its predicted uncertainty $\sigma$ follows the same per-dimension pattern as the empirical spread, which is what the constraint requires, since only the ratios of the predicted variances set the per-dimension weighting.

\paragraph{Dimension-wise constraint allocation}
Figure~\ref{fig:constraint_alloc} then shows how the learned uncertainty reshapes policy optimization, comparing the dimension-wise Mahalanobis constraint used by \ours{} with the state-wise constraint, an otherwise-identical variant of comparable overall tightness. The state-wise form applies a single per-state constraint bound and measures deviation by unweighted squared error, so each dimension's share follows the raw magnitude of $(\mathbf{a}_\theta-\hat{\mathbf{a}}^H_t)^2$: the largest share by far goes to $q_5$, one of the least consistently corrected dimensions. Normalizing by the predicted variances redistributes this budget: under the Mahalanobis form the largest share moves to $q_2$, the most consistently corrected dimension, the share of $q_5$ drops by more than half, and the most variable dimension $q_7$ receives the smallest share of all. This is exactly the design intent: along $q_2$ the correction constraint dominates and pulls the Residual Policy tightly toward the corrected action, whereas along the variable dimensions it relaxes and value maximization together with the VLA anchor loss drive the update. The state-level complement of this dimension-wise mechanism, the training dynamics of the state-dependent multiplier $\lambda_\omega(\mathbf{s}_t)$, is shown in Figure~\ref{fig:lambda_dynamics} of Appendix~\ref{sec:supp_lambda}.

\begin{figure}[t]
\centering
\resizebox{0.85\columnwidth}{!}{%
  \includegraphics{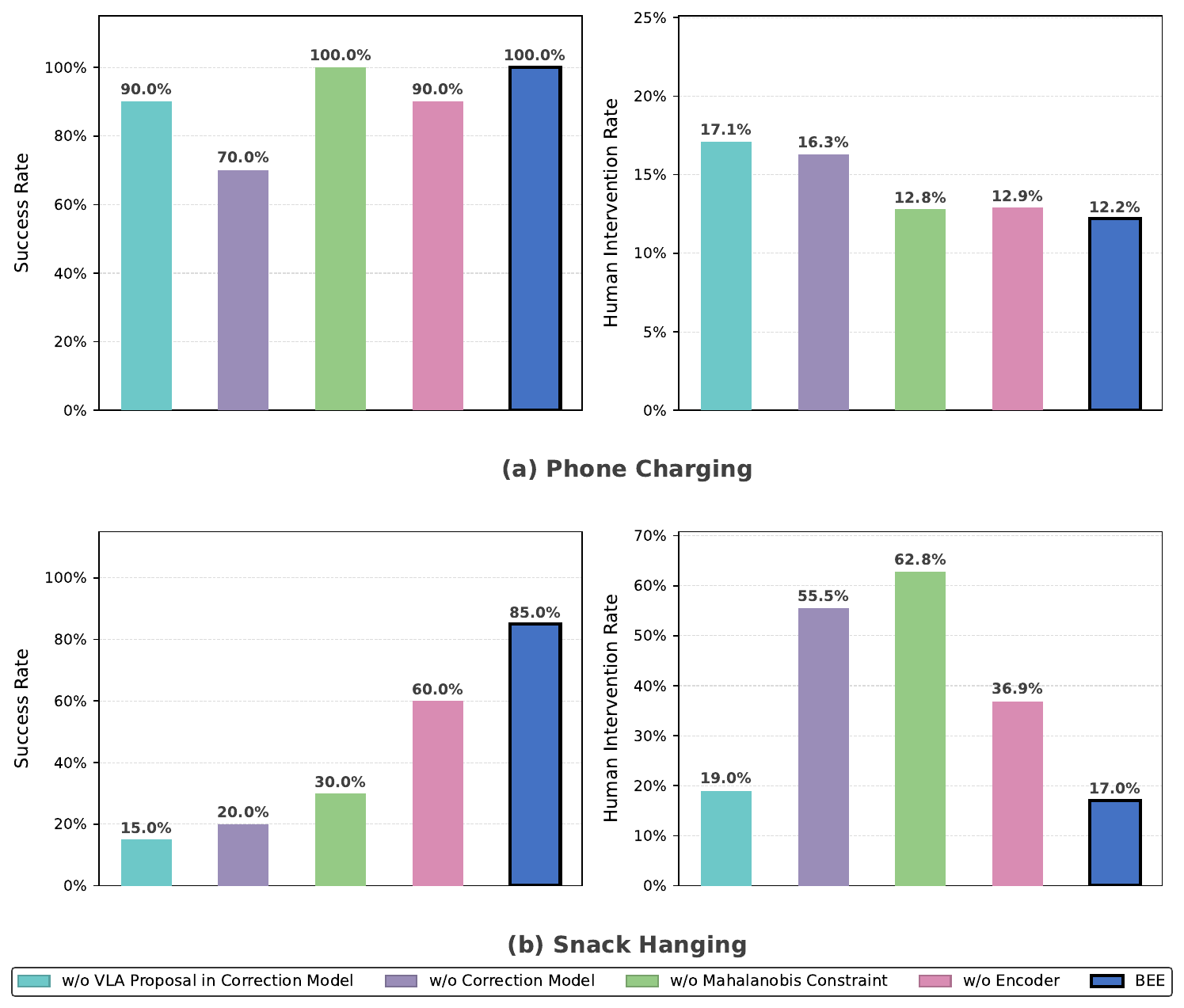}%
}
\caption{\textbf{Ablation of Human Correction Modeling} on \textbf{(a)}~phone charging and \textbf{(b)}~snack hanging, each reporting success rate (left) and human intervention rate (right) for \emph{w/o VLA Proposal in Correction Model} (the Correction Model conditioned on $s_t$ alone), \emph{w/o Correction Model} (direct actor supervision from raw human corrections), \emph{w/o Mahalanobis Constraint} (a single state-wise constraint in place of the dimension-wise one), \emph{w/o Encoder} (mean-pooled frozen VLA features in place of the latent-token encoder), and \ours{}.}
\label{fig:ablation_human}
\end{figure}

\subsection{Ablations}
\label{sec:exp_ablations}

Figure~\ref{fig:ablation_human} reports ablations of \ours{} on phone charging
and snack hanging. The two variants that touch the Correction Model are the
weakest: they hold the two lowest success rates on snack hanging and the two
highest intervention rates on phone charging, where the saturated task leaves
little headroom in success rate. \emph{w/o Correction Model} supervises the
actor with raw corrections, which are too inconsistent to serve as policy
targets. The gap is largest on snack hanging, where the Base Policy starts from
a near-zero success rate. \emph{w/o VLA Proposal in Correction Model} is weaker
still on that task: because the correction residual is defined
relative to the proposal it overrides, conditioning on the state alone leaves
the constraint center carrying the sampling deviation of the proposal instead
of cancelling it, and lets the predicted variances rank dimensions by how
variable the VLA is rather than by how consistently humans correct them. On
snack hanging, modeling corrections without the proposal is thus worse than not
modeling them at all. The constraint geometry matters as much: \emph{w/o
Mahalanobis Constraint} matches \ours{} on the saturated phone charging task
but falls far short on snack hanging, where the dimension-wise constraint
sharply raises success and cuts the intervention rate---the per-dimension
geometry is decisive exactly where a single scalar is not. Finally, \emph{w/o
Encoder} is the strongest of the ablated variants on snack hanging and stays
close to \ours{} on phone charging, so the gains stem from the proposal-conditioned
Correction Model and the dimension-wise constraint rather than from the token
interface, and \ours{} remains strong even without training a dedicated encoder.

\section{Conclusions}
\label{sec:conclusions}

We presented \ours{}, an intervention-adaptive framework for VLA-anchored real-world RL that converts human corrections of VLA proposals into an uncertainty-aware constraint on policy optimization, allowing the policy to move beyond expert imitation. At a matched online-data budget on three real-world manipulation tasks and a LIBERO-Pro simulation task, \ours{} attains the highest success rate on every task, 91.2\% on average against 57.5\% for RLT and 42.1\% for DSRL, and the lowest human intervention rate on all real-world tasks. Our ablations show that both the \emph{Correction Model} and the dimension-wise geometry of the constraint are necessary, and our analysis confirms that the learned uncertainty shifts the constraint onto the dimensions humans correct consistently. However, the predicted uncertainty is less reliable where human corrections are sparse, and extending to multi-phase tasks, multiple operators, and broader embodiments is a natural next step. We see modeling the structure of human corrections, rather than imitating them, as a promising direction for turning scarce real-world supervision into a reusable prior for VLA-anchored RL.

\newpage
\bibliographystyle{IEEEtran}
\bibliography{refs}

\newpage
\appendix
    
    \subsection{Task and Environment Details}
    \label{sec:supp_tasks}
    
    The main paper introduces the three real-world manipulation tasks, phone
    charging, snack hanging, and cloth aligning, and the simulated bowl placing task
    in LIBERO-Pro~\cite{liberopro}. Table~\ref{tab:supp_task_specs} summarizes how
    the four differ, and the subsections below give the platform, the action spaces,
    and the details of the simulated task.

    \begin{figure*}[!t]
    \centering
    \includegraphics[width=0.95\textwidth]{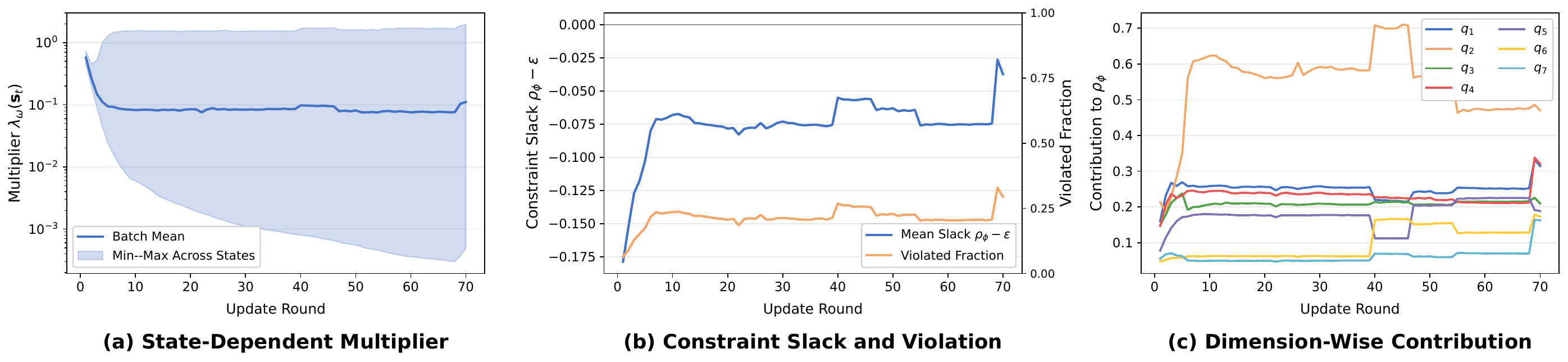}
    \caption{\textbf{Training Dynamics of the State-Dependent Multiplier} on phone charging.
    \textbf{(a)} State-wise multiplier statistics.
    \textbf{(b)} Constraint slack and violation rate.
    \textbf{(c)} Per-dimension contributions to $\rho_\phi$.}
    \label{fig:lambda_dynamics}
    \end{figure*}
    
    \begin{table*}[t]
    \centering
    \small
    \begin{tabular}{lcccc}
    \toprule
    & \textbf{Phone charging} & \textbf{Snack hanging} & \textbf{Cloth aligning}
    & \textbf{Bowl placing} \\
    \midrule
    Environment & real & real & real & LIBERO-Pro \\
    Control rate & $30$\,Hz & $30$\,Hz & $30$\,Hz & $20$\,Hz \\
    Action dims $d$ / $D$
    & $16$ / $160$ & $8$ / $160$
    & $16$ / $160$ & $7$ / $35$ \\
    VLA chunk length (approach) & $50$ & $50$ & $50$ & $10$ \\
    Learner chunk length $K$ & $10$ & $20$ & $10$ & $5$ \\
    Trained by online RL & critical phase & full task & critical phase
    & full task \\
    Success reported on & critical phase & full task & critical phase
    & full task \\
    Reward & sparse completion & sparse completion & dense, SAM~3
    & sparse completion \\
    Completion signal & operator button & operator button & mask IoU
    & operator button \\
    Success criterion & charger inserted & hung on rack & mask IoU $>0.85$
    & bowl on plate \\
    Online episode budget & $70$ & $70$ & $70$ & $25$ \\
    \bottomrule
    \end{tabular}
    \caption{\textbf{Per-Task Specification.} The action space lists the
    per-timestep dimension $d$ and the chunked dimension $D=Kd$ seen by the actor;
    the subset of dimensions online RL is allowed to edit, with the remaining
    dimensions held at the VLA proposal, is described in the text. On phone charging
    and cloth aligning online RL refines the
    precision-critical phase only, so an online episode covers that phase and success
    is reported on it. The online episode budget is the one given to the methods that
    are seeded with pre-collected human corrections. Section~\ref{sec:supp_protocol}
    gives the budget of the methods that are not.}
    \label{tab:supp_task_specs}
    \end{table*}
    
    \subsubsection{Robot Platform, Observations, and Actions}
    
    \paragraph{Platform and observations}
    The real-world tasks run on a dual-arm robot. Three $480\times640$ RGB cameras
    observe the workspace, one on the head and one
    on each wrist. The frozen VLA consumes these views as $\mathbf{o}_t$, the language
    instruction $g$, and the proprioceptive state $\mathbf{p}_t$, which holds the arm
    joint states and the effector states, whereas the online learner observes only the
    latent token $\mathbf{z}_t$ produced by the frozen extractor together with
    $\mathbf{p}_t$. The dense reward for cloth aligning is computed from the head-camera
    view.
    
    \paragraph{Control frequency and action space}
    Table~\ref{tab:supp_task_specs} gives the per-task values, which differ in both
    the action representation and the learner chunk length. Phone charging and cloth
    aligning are the two dual-arm tasks and are controlled in joint space, as
    fourteen dual-arm joint dimensions together with two further end-effector
    dimensions, and online RL is allowed to edit only the joint dimensions of the
    right arm, which performs the precision-critical step and is also the arm the
    operator intervenes on. Snack hanging and bowl placing are instead controlled by
    end-effector commands, and online RL edits the pose dimensions while the gripper
    is held at the VLA proposal.

    \subsubsection{Simulation Task: Bowl Placing}
    
    Bowl placing is task $9$ of the \texttt{libero\_spatial} suite of
    LIBERO-Pro~\cite{liberopro}, an evaluation extension of the LIBERO
    benchmark~\cite{libero}.

    \subsection{Reward Functions}
    \label{sec:supp_reward}
    
     Phone charging,
    snack hanging, and bowl placing use a sparse task-completion reward, and cloth
    aligning uses a vision-based dense reward.
    
    \paragraph{Sparse task-completion reward}
    For phone charging, snack hanging, and bowl placing, the per-step reward is
    $r_t=0$ throughout the episode and $r_t=1$ only when the task is judged complete,
    after which the episode terminates. 
    
    \paragraph{Vision-based dense reward for cloth aligning}
    We derive the reward from vision, as the increment of an
    alignment potential between the cloth and the target region. The same perception
    module also defines the success criterion for this task, since a trial counts as a
    success when the intersection over union of the cloth mask and the target-region
    mask exceeds $0.85$ at some point in the episode.
    
    \subsection{Implementation Details}
    \label{sec:supp_impl}
    
    Table~\ref{tab:supp_hyperparams} gives the hyperparameters of the phone-charging configuration, the run analyzed throughout the main paper; the per-task control rates, action dimensions, chunk lengths, and online budgets are listed in Table~\ref{tab:supp_task_specs}. The paragraphs below situate those values against the method of the main paper.
    
    \paragraph{Backbone and frozen interface}
    All experiments start from the open-source $\pi_{0.5}$ weights~\cite{pi05}, fine-tuned with behavior cloning on each task separately and then frozen, so the frozen VLA is task-specific. The extractor that maps its internal features to the latent token $\mathbf{z}_t$ is an RLT-style latent-token encoder~\cite{rlt}, a four-layer encoder and four-layer decoder of width $2048$ with $16$ attention heads, so $\mathbf{z}_t$ has dimension $2048$. It is frozen as well, so the interface pair $(\mathbf{s}_t,\tilde{\mathbf{a}}_t)$ stays fixed throughout online training.

    \begin{table*}[t]
    \centering
    \footnotesize
    \begin{tabular}[t]{ll}
    \toprule
    Hyperparameter & Value \\
    \midrule
    \multicolumn{2}{l}{\emph{Interface and action space}} \\
    Control frequency & 30\,Hz \\
    VLA chunk length, approach / online & 50 / 10 \\
    Learner chunk length $K$ & 10 \\
    Per-timestep action dim.\ $d$ & 16 \\
    Chunked action dim.\ $D=Kd$ & 160 \\
    RL-editable dimensions & 7 arm joints \\
    Action normalization & $\pi_{0.5}$ z-score \\
    \midrule
    \multicolumn{2}{l}{\emph{Networks and optimization}} \\
    MLP depth / width (all modules) & 3 / 512 \\
    Critic heads & 2, min over heads \\
    Discount $\gamma$ & 0.99 \\
    Target update rate $\tau$ & 0.005 \\
    Update-to-data ratio (critic, policy) & 5 \\
    Critic updates per policy update & 2 \\
    Batch size & 256 \\
    Optimizer & Adam($0.9$, $0.95$, $\epsilon{=}10^{-8}$) \\
    Learning rate ($\theta$, $\psi$, $\phi$) & $3\times10^{-4}$ \\
    Multiplier learning rate ($\omega$) & $3\times10^{-6}$ \\
    Gradient-norm clip & 1.0 \\
    Exploration noise $\sigma_{\mathrm{expl}}$ & 0.01 \\
    \bottomrule
    \end{tabular}
    \hspace{0.03\textwidth}
    \begin{tabular}[t]{ll}
    \toprule
    Hyperparameter & Value \\
    \midrule
    Correction Model batch size & 256 \\
    Update interval $M$ (samples) & 100 \\
    Update-to-data ratio $N$ & 0.12 \\
    Constraint bound $\varepsilon$ & 0.3 \\
    Per-dimension variance floor & 0.02 \\
    VLA anchor weight $\eta$ & 1.0 \\
    \midrule
    \multicolumn{2}{l}{\emph{Data and exploration}} \\
    Seed episodes in $\mathcal{B}_H$ & $20$ \\
    $\mathcal{B}_H$-to-$\mathcal{B}_R$ sampling ratio & 1:1 \\
    Reference-action dropout & 0.5 \\
    VLA warmup before handover & 5 weight syncs \\
    \bottomrule
    \end{tabular}
    \caption{\textbf{Hyperparameters of \ours{} on Phone Charging.} Values are those of the run analyzed in the main paper; the quantities that vary by task are listed per task in Table~\ref{tab:supp_task_specs}. All distances, thresholds, and noise scales are expressed in normalized action units.}
    \label{tab:supp_hyperparams}
    \end{table*}
    
    \paragraph{Hardware and compute}
    Evaluation overlaps inference with execution by real-time chunking, on the platform described in Section~\ref{sec:supp_tasks}. Over its episode budget, one online run of \ours{} performs between roughly $20{,}000$ and $90{,}000$ gradient updates depending on the task, since the update-to-data ratio is fixed while the episode length and the online budget are not. Training length is measured in online episodes throughout rather than in wall-clock time, so a run ends when its episode budget is exhausted. The learner and VLA inference run on a single RTX~4090D for every method except DAgger, which was trained on an A100.
    
    \subsection{Intervention Protocol and Evaluation Details}
    \label{sec:supp_protocol}
    
    \paragraph{Matched robot-data budget}
    Every method is given the same total amount of interaction with the robot, which
    is $90$ episodes on each of the three real-world tasks and $45$ on bowl placing.
    What differs is how that total is split between episodes collected before online
    training and episodes collected during it. \ours{}, RLT, SiLRI, and HIL-SERL
    maintain a demonstration buffer, so around $20$ of their episodes are collected
    in advance as human corrections and seed $\mathcal{B}_H$, leaving the online
    budget of $70$ episodes on the real-world tasks and $25$ on bowl placing that
    Table~\ref{tab:supp_task_specs} lists. DSRL and DAgger maintain no demonstration
    buffer, and seeding them with one would depart from their original formulations,
    since both already start from the same fine-tuned VLA that was trained on task
    demonstrations. They therefore spend those $20$ episodes online instead, for $90$
    online episodes on the real-world tasks and $45$ on bowl placing. Reported
    success rates are read at the end of this budget for every method.
    
    \paragraph{Intervention protocol}
    During online training a human operator monitors execution and takes over
    whenever the robot is about to fail in a way it cannot recover from on its own,
    which in practice concentrates the corrections on the precision-critical contact
    phase of each task. In
    simulation the operator instead uses a $6$-DoF SpaceMouse, with two buttons for
    the gripper.
    
    \paragraph{Phase switching}
    For the two-phase tasks, phone charging and cloth aligning, RLT and \ours{}
    switch from the frozen VLA approach to online RL using a progress head trained
    during SFT, entering the precision-critical phase once the predicted progress
    crosses a threshold.
    
    \subsection{Baseline Implementations}
    \label{sec:supp_baselines}
    
    Because the comparisons are conducted on a real robot, all baselines are
    implemented under a common setup. Every method is trained and evaluated on the
    same tasks with the same reward source, the same VR interface for human
    interventions and completion signals, the same action normalization, and the
    same matched robot-data budget and evaluation protocol described above. The
    methods that build on a frozen VLA start from the identical fine-tuned
    $\pi_{0.5}$ checkpoint and, where they use a latent-token interface, from the
    identical frozen extractor. SiLRI and HIL-SERL are the exception, since their
    original formulations do not assume a pretrained VLA, and we retain that setting
    rather than reworking them into ours. Following their published implementations,
    both encode the camera views with a ResNet-10 initialized from the publicly
    released pretrained weights and train the policy and the critic on top of it, so
    what is trained from scratch in these two baselines is the policy rather than the
    visual representation. The paragraphs below give the per-method details.

    \begin{table}[!t]
    \centering
    \begin{tabular}{lcc}
    \toprule
    Method & SR (\%)\,$\uparrow$ & Int.\ (\%)\,$\downarrow$ \\
    \midrule
    \multicolumn{3}{l}{\emph{Phone charging}} \\
    Base Policy & $90.0 \pm 5.0$ & -- \\
    DAgger   & $63.3 \pm 5.8$ & $7.8$ \\
    \ours{}  & $100.0 \pm 0.0$ & $12.2$ \\
    \midrule
    \multicolumn{3}{l}{\emph{Bowl placing}} \\
    SiLRI    & $0.0$ & $90.0+$ \\
    HIL-SERL & $0.0$ & $90.0+$ \\
    \ours{}  & $90.0 \pm 13.2$ & $27.4$ \\
    \midrule
    \multicolumn{3}{l}{\emph{Snack hanging}} \\
    SiLRI    & $0.0$ & $90.0+$ \\
    HIL-SERL & $0.0$ & $90.0+$ \\
    \ours{}  & $85.0 \pm 0.0$ & $17.0$ \\
    \bottomrule
    \end{tabular}
    \caption{\textbf{Comparison with DAgger, SiLRI, and HIL-SERL} under the matched
    robot-data budget. Each method is evaluated on the task where the comparison is
    informative. Intervention rates follow the accounting in
    Section~\ref{sec:supp_protocol} and are not directly comparable across methods.}
    \label{tab:supp_more_baselines}
    \end{table}
    
    \paragraph{Base Policy}
    The number of demonstrations used for supervised fine-tuning differs by task and
    ranges from $200$ to $900$ episodes, so the online budget every method is then
    given is a small fraction of the data the shared starting point was trained on.
    The Base Policy consumes none of that budget and is also the initialization of
    the frozen-VLA methods.
    
    \paragraph{RLT}
    RLT~\cite{rlt} has no public implementation and is reimplemented on the shared
    chunk-level actor--critic backbone, so it uses the same frozen VLA interface and
    the same action envelope as \ours{} and the comparison isolates how the two
    methods use human corrections. Following the original RLT rollout procedure,
    when the human intervenes, the human correction replaces the VLA proposal
    stored for that transition.
    
    \paragraph{DSRL}
    DSRL~\cite{dsrl} acts on the same frozen fine-tuned VLA and is run in the
    chunk-$10$ configuration of the original formulation: noise is sampled for the
    first $10$ steps of a chunk, the last noise vector is repeated to fill the
    $50$-step VLA chunk, and the first ten actions of the result are executed.
    
    \paragraph{DAgger}
    DAgger~\cite{dagger} is applicable on top of the same fine-tuned VLA, and we
    evaluate it on phone charging, where its interactive-expert assumption applies
    directly. It lands below both \ours{} and the Base Policy it starts from, even
    though it receives $20$ additional online episodes in place of the seed
    corrections, so aggregating and imitating the operator's corrections degrades the
    very policy it was meant to improve.
    
    \paragraph{SiLRI and HIL-SERL}
    Neither is competent enough at this budget to be left running, so the operator
    holds control for almost the entire episode throughout, and neither run makes
    progress without that continuous intervention.
    
    \paragraph{Scope of the comparison}
    Table~\ref{tab:supp_more_baselines} compares \ours{} with DAgger, SiLRI, and
    HIL-SERL under the same intervention protocol and robot-data budget as in the
    main paper. Each is evaluated on the task where the comparison is informative.
    SiLRI and HIL-SERL train task-specific visuomotor policies from scratch, and all
    of our tasks span roughly a minute of continuous control, so a budget of tens of
    episodes is far below what a from-scratch policy needs on any of them.
    
    \subsection{Training Dynamics of the State-Dependent Multiplier}
    \label{sec:supp_lambda}
    
    The main paper analyzes how the dimension-wise constraint allocates its budget across action dimensions. Figure~\ref{fig:lambda_dynamics} complements that snapshot with the state-level behavior of the constraint during training, and three readings matter. The multiplier does not converge to one value: it stays near zero on states where the policy already agrees with the corrected action and large on the subset of states where the constraint is active, so a single global multiplier would have to average over two regimes that the state-dependent parameterization keeps apart. The violation rate settles well above zero rather than vanishing, which is what an interactive setting should produce, since newly collected corrections keep introducing states where the policy deviates from the corrected action and the multiplier re-tightens locally on them. And the per-dimension decomposition of $\rho_\phi$ reproduces the allocation the main paper reads off a single snapshot, so that mechanism persists throughout online training rather than holding at one point in it.

\end{document}